\documentclass{article}

\PassOptionsToPackage{numbers, compress}{natbib}
\usepackage[final]{neurips_2023}
\usepackage{graphicx}

\usepackage[utf8]{inputenc} 
\usepackage[T1]{fontenc}    
\usepackage{hyperref}       
\usepackage{url}            
\usepackage{booktabs}       
\usepackage{amsfonts}       
\usepackage{nicefrac}       
\usepackage{microtype}      
\usepackage{xcolor}         

\title{Mathify: Evaluating Large Language Models on Mathematical Problem Solving Tasks}

\author{%
  Avinash Anand \\
  MIDAS Labs\\
  IIIT-Delhi, Delhi, India \\
  \texttt{avinasha@iiitd.ac.in} \\
  \And
  Mohit Gupta \\
  MIDAS Labs\\
  IIIT-Delhi, Delhi, India \\
  \texttt{mohit22112@iiitd.ac.in} \\
  \AND
  Kritarth Prasad \\
  MIDAS Labs\\
  IIIT-Delhi, Delhi, India \\
  \texttt{kritarth20384@iiitd.ac.in} \\
  \And
  Navya Singla \\
  MIDAS Labs\\
  IIIT-Delhi, Delhi, India \\
  \texttt{singlanavya01@gmail.com} \\
  \And
  Sanjana Sanjeev \\
  MIDAS Labs\\
  IIIT-Delhi, Delhi, India \\
  \texttt{sanjana21094@iiitd.ac.in} \\
  \And
  Jatin Kumar \\
  MIDAS Labs\\
  IIIT-Delhi, Delhi, India \\
  \texttt{jatin20206@iiitd.ac.in} \\
  \And
  Adarsh Raj Shivam \\
  MIDAS Labs\\
  IIIT-Delhi, Delhi, India \\
  \texttt{adarsh20274@iiitd.ac.in} \\
  \And
  Rajiv Ratn Shah \\
  MIDAS Labs\\
  IIIT-Delhi, Delhi, India \\
  \texttt{rajivratn@iiitd.ac.in} \\
}

\begin{document}
\maketitle
\begin{abstract}
The rapid progress in the field of natural language processing (NLP) systems and the expansion of large language models (LLMs) have opened up numerous opportunities in the field of education and instructional methods. These advancements offer the potential for tailored learning experiences and immediate feedback, all delivered through accessible and cost-effective services. One notable application area for this technological advancement is in the realm of solving mathematical problems. Mathematical problem-solving not only requires the ability to decipher complex problem statements but also the skill to perform precise arithmetic calculations at each step of the problem-solving process. However, the evaluation of the arithmetic capabilities of large language models remains an area that has received relatively little attention. In response, we introduce an extensive mathematics dataset called "\textbf{MathQuest}" sourced from the 11th and 12th standard Mathematics NCERT textbooks. This dataset encompasses mathematical challenges of varying complexity and covers a wide range of mathematical concepts. Utilizing this dataset, we conduct fine-tuning experiments with three prominent LLMs: LLaMA-2, WizardMath, and MAmmoTH. These fine-tuned models serve as benchmarks for evaluating their performance on our dataset. Our experiments reveal that among the three models, \textbf{MAmmoTH-13B} emerges as the most proficient, achieving the highest level of competence in solving the presented mathematical problems. Consequently, \textbf{MAmmoTH-13B} establishes itself as a robust and dependable benchmark for addressing NCERT mathematics problems. GitHub repository: \href{https://github.com/midas-research/mathify}{\textbf{https://github.com/midas-research/mathify}}.
\end{abstract}

\section{Introduction}
Mathematical problem-solving represents a multifaceted cognitive skill, encompassing the comprehension of problem statements, identification of pertinent concepts and formulas, application of suitable strategies and algorithms, precise calculations, and the verification of solution validity and reasonableness. Traditionally, mathematical problem-solving has been imparted and assessed through conventional means such as textbooks, worksheets, and examinations, often affording limited feedback and learner guidance. Furthermore, these methods may not fully capture the diversity and intricacy of real-world mathematical challenges encountered by students.

In the era of rapid advancements in artificial intelligence and natural language processing (NLP), large language models (LLMs) have emerged as formidable tools for generating natural language text across a spectrum of domains and tasks~\cite{unknown}. LLMs, grounded in the transformer architecture~\cite{vaswani2023attention}, have the capacity to glean long-range dependencies and contextual representations from vast corpora of text data. These LLMs have showcased impressive proficiency in mathematical reasoning and problem-solving by leveraging their inherent understanding of arithmetic operations, algebraic principles, and symbolic manipulation. Nevertheless, existing LLMs grapple with substantial hurdles in tackling math word problems, particularly those necessitating intricate reasoning, multi-step arithmetic calculations, or domain-specific knowledge ~\cite{heyueya2023solving,article1,wu2023empirical}.

\begin{figure}[ht]
\centering
  \includegraphics[width=0.8\linewidth]{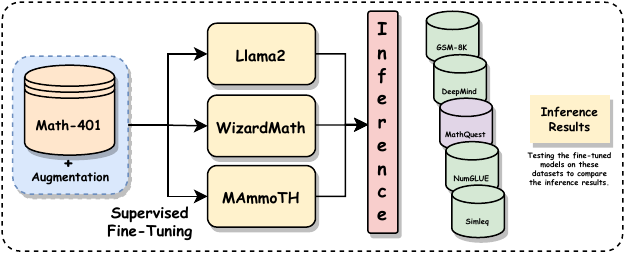}
  \caption{This figure shows the fine-tuning flow, the LLMs we use for fine-tuning, and the datasets we use for inference.} 
\label{fine_tuning_flow}
\end{figure}

The advent of large language models (LLMs) has proven to be a boon in the field of education, as evidenced by recent studies~\cite{milano2023large,olga2023generative,yan2023practical}. These versatile models have ushered in a new era of learning possibilities, catering to individual student needs by considering their preferences, objectives, interests, and aptitudes. For instance, LLMs offer a tailored learning experience, providing personalized feedback, guidance, explanations, and recommendations~\cite{Jeon_Lee_2023}. Educators, too, find these models invaluable, as they simplify the creation of engaging learning materials such as quizzes, summaries, questions, and exercises~\cite{moore2023empowering}. Notably, LLMs can even generate multiple-choice questions based on provided text passages. Additionally, these models excel in enhancing language proficiency, aiding learners in vocabulary, grammar, pronunciation, and fluency~\cite{Jeon_Lee_2023}. Their versatility extends to assisting students and researchers in exploring new topics and extracting information from diverse sources. They effortlessly generate summaries~\cite{xiao2023enhancing}, identify keywords, generate citations~\cite{citation_text_generation,10.1007/978-3-031-49601-1_3,inbook}, and provide relevant links in response to queries.

This paper endeavors to tackle the challenges posed by mathematical problem-solving within the context of LLMs. To this end, we introduce MathQuest, a comprehensive mathematics dataset meticulously curated from the 11th and 12th standard Mathematics NCERT textbooks\footnote{\url{https://ncert.nic.in/}}. This dataset spans various levels of mathematical complexity and encompasses a wide array of mathematical concepts. We introduce this dataset because existing open-source datasets primarily consist of relatively straightforward mathematical problems. In contrast, standard mathematical problems can be significantly more complex. To equip Large Language Models (LLMs) with the ability to solve these intricate problems, we conduct fine-tuning on this dataset. Furthermore, we propose a novel approach for fine-tuning three preeminent LLMs: MAmmoTH~\cite{yue2023mammoth}, LLaMA-2~\cite{touvron2023llama}, and WizardMath~\cite{luo2023wizardmath} using our MathQuest dataset. Our evaluation encompasses not only the performance of these fine-tuned models on our dataset but also their proficiency on other openly accessible mathematical reasoning datasets. Our findings indicate that \textbf{MAmmoTH-13B} outshines its counterparts, emerging as the most adept and proficient in solving the mathematical challenges presented. Thus, \textbf{MAmmoTH-13B} establishes itself as a dependable and robust baseline for addressing NCERT mathematics problems.

\section{Related Work\label{related_work}}
In this section, we delve into the existing literature, unveiling a diverse array of approaches utilizing Large Language Models (LLMs) for tackling mathematical problems.

Recent research has highlighted the potential of Large Language Models (LLMs) in education~\cite{10.1007/978-3-031-49601-1_4,10.1007/978-3-031-49601-1_5}. They offer promise in automating question generation and supporting direct interactions within the learning environment~\cite{KASNECI2023102274}. Furthermore, investigations have explored few-shot prompting techniques over LLMs for addressing mathematical word problems~\cite{wei2023chainofthought,zhou2023leasttomost,gao2023pal}. The "chain-of-thought" prompting approach~\cite{wei2023chainofthought} leverages explicit intermediate reasoning steps to bolster the LLM's reasoning abilities. To mitigate arithmetic errors commonly observed in LLMs~\cite{lewkowycz2022solving,hendrycks2021measuring}, earlier studies~\cite{chowdhery2022palm} have explored the use of external calculators to execute operations generated by LLMs.

\begin{figure}[ht]
\centering
  \includegraphics[width=7cm, height=5cm]{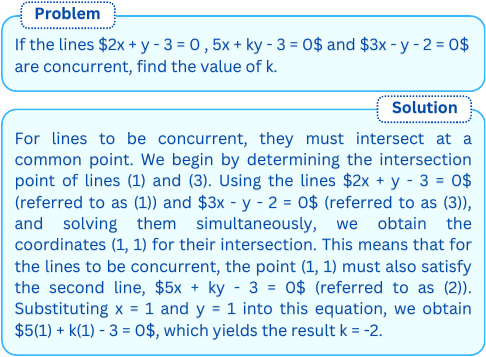}
  \caption{Our Dataset \textbf{MathQuest} Sample} 
\label{sample_data}
\end{figure}

Furthermore, \cite{weng2023large} presents a novel method tailored for addressing elementary arithmetic and logical problems. This method concatenates the generated answer with the original problem statement, tasking the model with predicting the initial conditions to verify the accuracy of the answer. Notably, a subset of these approaches~\cite{Drori_2022,chen2022program} can function effectively with zero-shot prompts, offering a versatile approach to mathematical problem-solving. A specialized method, MathPrompter~\cite{imani2023mathprompter}, targets the enhancement of arithmetic operations and reasoning capabilities of LLMs, particularly designed to facilitate mathematical problem-solving tasks.

Various approaches exist for enhancing mathematical problem-solving with Large Language Models (LLMs). Wang et al.'s self-consistency~\cite{wang2023selfconsistency}, built on the CoT framework, assesses multiple potential reasoning paths and selects answers via majority vote. ~\cite{li2023making} extend self-consistency by teaching a verifier to validate each step, while ~\cite{madaan2023selfrefine} use recent LLMs like GPT-3.5 to generate an output, provide feedback, and prompt the model for improvements. \cite{wang2021exploring} evaluate pretrained language models on basic arithmetic expressions, including addition $(+)$ and subtraction $(-)$, and \cite{muffo2023evaluating} expand the assessment to include multiplication $(*)$ operations within the language models' scope.

\section{Dataset\label{dataset}}
For our research experiments, we employed the Math-401 dataset~\cite{
yuan2023large}, which encompasses 401 samples of mathematical problems. This dataset encompasses a diverse range of mathematical operations, including addition $(+)$, subtraction $(-)$, multiplication $(*)$, division $(/)$, exponentiation, trigonometric functions ($\sin, \cos, \tan$), logarithmic functions $(\log, \ln)$, and incorporates integers, decimals, and irrational numbers ($\pi, e$). Recognizing the limited sample size of this dataset for effective learning by large language models, we expanded it through augmentation, resulting in a dataset size of $302,000$ samples. To construct our augmented dataset, we employed the \textbf{\textit{SymPy}} Python library. This library allowed us to generate arithmetic mathematical equations along with their corresponding ground truth values. These equations covered basic arithmetic operators such as addition (+), subtraction (-), multiplication (*), and division (/). Furthermore, the dataset includes extensive arithmetic expressions with brackets, mimicking the complexity often encountered in real-world math word problems. Table~\ref{augmented_dataset_distribution} provides a comprehensive breakdown of the question types utilized in the creation of our augmented dataset. Furthermore, we evaluated our model's performance on four additional datasets: GSM-8K~\cite{cobbe2021gsm8k}, DeepMind~\cite{saxton2019analysing}, NumGLUE~\cite{mishra-etal-2022-numglue}, and SimulEq~\cite{koncel-kedziorski-etal-2016-mawps}.

\begin{table}[ht]
    \centering
    \renewcommand{\arraystretch}{1.2}
    \setlength{\tabcolsep}{1.4\tabcolsep}
    \begin{tabular}{|@{}c@{}|@{}c@{}|@{}c@{}|}
    \hline
        \begin{tabular}{r|r|r|r|r}
           \small \textbf{Type} & \small \textbf{Range} & \small \textbf{Decimal Places (1 - 4)} & \small \textbf{Variables} & \small \textbf{Count} \\\hline\hline
           Small Integer & [-20, 20] & $\times$ & (x, y) & 65,000 \\\hline
           Small Decimal & [-20, 20] & \checkmark & (x, y) & 35,000 \\\hline
           Small Decimal + Integer & [-20, 20] & \checkmark & (x, y) & 39,000 \\\hline
           Large Integer & [-1000, 1000] & $\times$ & (x, y) & 39,000 \\\hline
           Large Decimal & [-1000, 1000] & \checkmark & (x, y) & 25,000 \\\hline
           Large Decimal + Integer & [-1000, 1000] & \checkmark & (x, y) & 25,000 \\\hline
           3 Terms & [-100, 100] & \checkmark & (x, y, z) & 25,000 \\\hline
           4 Terms & [-100, 100] & \checkmark & (w, x, y, z) & 49,000 \\\hline
           \textbf{Total} &-&-&-& 302,000 \\\hline
        \hline
        \end{tabular}
    \end{tabular}
    \vspace{0.3cm}
    \caption{The distribution of types of question in our augmented Math-401 dataset}
\label{augmented_dataset_distribution}
\end{table}

\subsection{Our Dataset: MathQuest}
\begin{figure}[ht]
\centering
  \includegraphics[width=12cm, height=9cm]{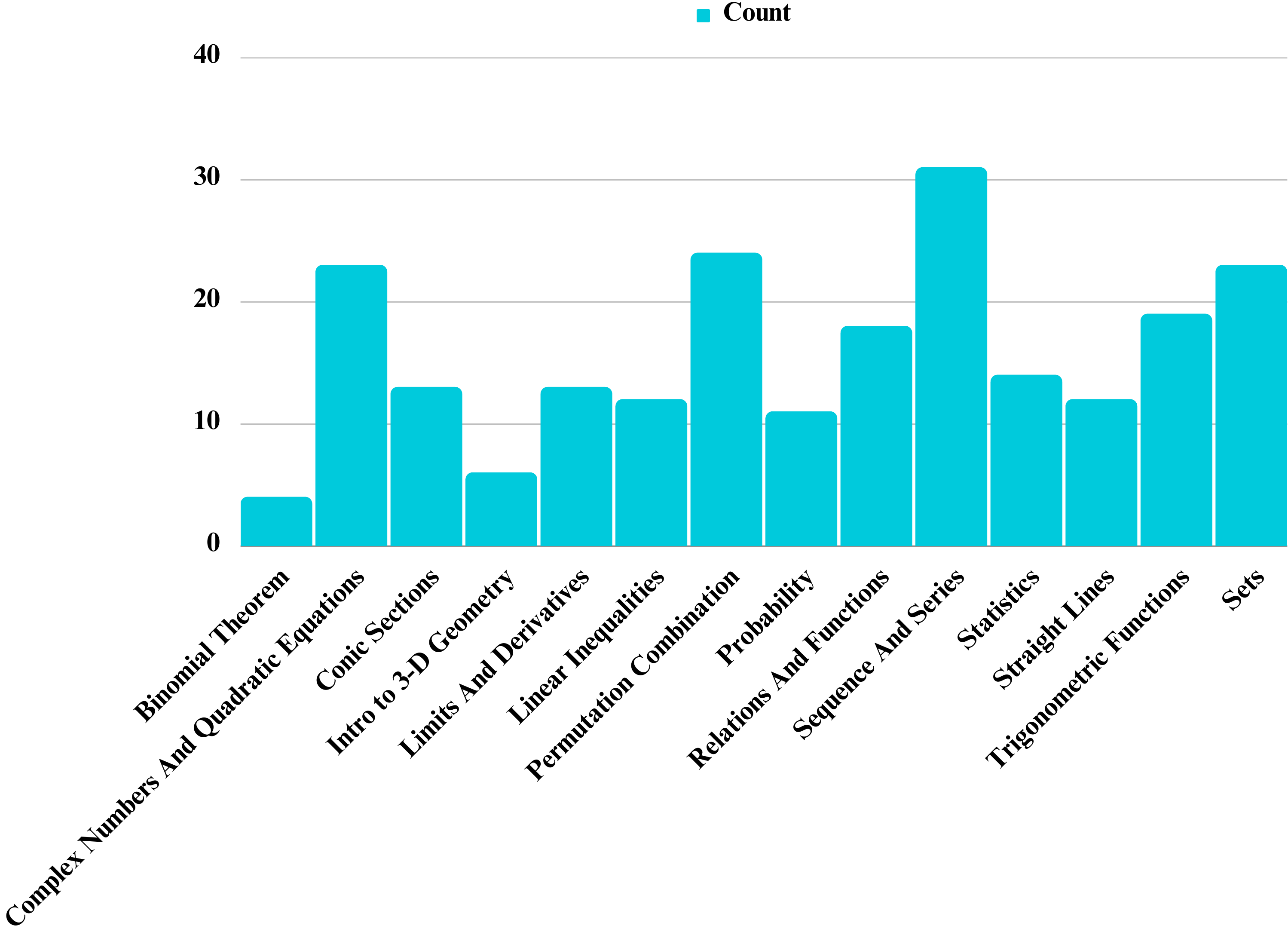}
  \caption{Distribution of Count of Samples of each Concept} 
\label{data_distribution}
\end{figure}

We have meticulously curated our own dataset, referred to as \textbf{MathQuest}, sourcing problems from high school mathematics NCERT books. MathQuest is a rich resource, encompassing word problems of varying complexities and spanning diverse mathematical concepts. Our dataset comprises a total of 14 overarching mathematical domains, including sets, trigonometry, binomial theorem, and more. The distribution of samples across these concepts is visually represented in Figure.\ref{data_distribution}. Our dataset contains total of 223 samples. Notably, as depicted in the charts, the category of "Sequence and Series" boasts the highest number of problems within our dataset. To provide a glimpse of our dataset's structure, we present a sample from MathQuest in Figure.\ref{sample_data}.

\section{Methodology\label{methodology}}
This research aims to enhance the mathematical problem-solving capabilities of large language models. Initially, we observed that existing open-source models such as LLaMA-2~\cite{touvron2023llama} and Vicuna~\cite{vicuna2023} struggled with elementary mathematical tasks like simple addition and subtraction. This observation served as the catalyst for our research, motivating us to improve LLMs' proficiency in comprehending and accurately solving mathematical problems.

To achieve this, we adopted a instructive approach reminiscent of teaching mathematics to students. We commenced by imparting a clear understanding of fundamental operators such as $+, -, *, /$, gradually progressing to more advanced operators and expressions. Similarly, we endeavored to acquaint LLMs with the meanings of mathematical operators and expressions. To facilitate this process, we leveraged the Math-401 dataset~\cite{yuan2023large}, a valuable resource comprising 401 data samples consisting of basic mathematical questions and their corresponding answers. Given the dataset's limited size, we augmented it to introduce greater diversity and complexity, ensuring that the model could grasp and master advanced mathematical concepts during training.

For the fine-tuning process, we employed three prominent large language models: LLaMA-2~\cite{touvron2023llama}, WizardMath~\cite{luo2023wizardmath}, and MAmmoTH~\cite{yue2023mammoth}. LLaMA-2~\cite{touvron2023llama} represents an upgraded version of LLaMA, refined through training on an enriched mixture of publicly available data. The enhancements encompass a 40\% increase in the pre-training corpus size, a doubling of the model's context length, and the incorporation of grouped-query attention.

WizardMath~\cite{luo2023wizardmath} introduces an innovative approach known as Reinforcement Learning from Evol-Instruct Feedback (RLEIF). This method combines Evol-Instruct and reinforced process supervision techniques to evolve GSM8k and MATH datasets. Subsequently, it fine-tunes the pre-trained LLaMA-2 model using the evolved data and reward models, resulting in the development of the WizardMath model.

Lastly, the MAmmoTH~\cite{yue2023mammoth} models are trained using the MathInstruct dataset, meticulously curated for instructional tuning. MathInstruct is constructed from a compilation of 13 mathematical datasets, including six newly curated rationales. It encompasses a hybrid of chain-of-thought (CoT) and program-of-thought (PoT) rationales, ensuring comprehensive coverage of diverse mathematical domains. The entire fine-tuning process is outlined in Figure.~\ref{fine_tuning_flow}.

\begin{table}[ht]
    \centering
    \renewcommand{\arraystretch}{1.2}
    \setlength{\tabcolsep}{0.6\tabcolsep}
    \begin{tabular}{|@{}c@{}|@{}c@{}|@{}c@{}|}
    \hline
        \begin{tabular}{l|l}
           \textbf{Model} & \small \textbf{\# of Params }\\\hline\hline\\\hline\hline
           \small \textbf{LLaMA-2} & 7B \\\hline
           \small \textbf{LLaMA-2} & 13B\\\hline
           \small \textbf{WizardMath} & 7B\\\hline
           \small \textbf{WizardMath} & 13B\\\hline
           \small \textbf{MAmmoTH} & 7B\\\hline
           \small \textbf{MAmmoTH} & 13B \\\hline\hline
        \end{tabular} & 
        \begin{tabular}{r|r|r|r|r|r}
           \multicolumn{6}{c}{\small \textbf{Accuracy }}\\\hline\hline
           \small \textbf{GSM-8K} & \small \textbf{DeepMind} & \small \textbf{NumGLUE} & \small \textbf{SimulEq} & \small \textbf{Math-401*} & \small \textbf{MathQuest} \\\hline\hline
           16.0 & 46.0 & 37.0 & 11.0 & 10.0 & 10.4 \\\hline
           22.0 & 50.0 & 42.0 & 15.0 & 10.0 & 14.1 \\\hline
           61.0 & 51.0 & 54.0 & 27.0 & 6.0 & 14.6 \\\hline
           65.0 & 55.0 & 70.0 & 36.0 & 8.0 & 14.3 \\\hline
           43.0 & 49.0 & 54.0 & 23.0 & 11.0 & 12.2 \\\hline
           44.0 & 48.0 & 56.0 & 26.0 & 14.0 & 18.1 \\\hline\hline
        \end{tabular}
    \end{tabular}
    \vspace{0.3cm}
    \caption{Exact Match Accuracy results on the set of 100 samples of 5 datasets and our dataset MathQuest \textbf{Before} fine-tuning on Math-401 dataset. (*) refers to the set of Math-401 we augmented for fine-tuning.}
\label{accuracy1}
\end{table}

\section{Experiments\label{experiments}}
In this section, we delve into the details of our conducted experiments, outlining the experimental setup and the utilized hyper-parameters. Our research objective revolves around the creation of a high school-level mathematical dataset, encompassing questions of varying complexities and diverse concepts, followed by the establishment of robust baselines for solving mathematical problems.

To achieve this, we conducted experiments involving three prominent large language models: LLaMA-2~\cite{touvron2023llama}, WizardMath~~\cite{yue2023mammoth}. We performed these experiments on both the 7B and 13B variants of these large language models (LLMs). Our experiments were executed in two stages. In the first stage, we directly loaded the original model weights and carried out inference on our designated test set. In the second stage, we undertook the fine-tuning of these models using the Math-401~\cite{yuan2023large} dataset as a crucial step in the process.

The Math-401~\cite{yuan2023large} dataset initially comprised 401 elementary mathematical equations paired with their corresponding results. To enhance its comprehensiveness and diversity, we performed data augmentation by introducing more intricate equations involving operators such as addition ($+$), subtraction ($-$), multiplication ($*$), division ($/$), as well as parentheses ($()$). This augmentation process aimed to create a more generalized and versatile dataset. Subsequently, we proceeded to fine-tune the Large Language Models (LLMs) using this augmented Math-401~\cite{yuan2023large} dataset.

\begin{table}[ht]
    \centering
    \renewcommand{\arraystretch}{1.2}
    \setlength{\tabcolsep}{0.6\tabcolsep}
    \begin{tabular}{|@{}c@{}|@{}c@{}|@{}c@{}|}
    \hline
        \begin{tabular}{l|l}
           \textbf{Model} & \small \textbf{\# of Params }\\\hline\hline\\\hline\hline
           \small \textbf{LLaMA-2} & 7B \\\hline
           \small \textbf{LLaMA-2} & 13B\\\hline
           \small \textbf{WizardMath} & 7B\\\hline
           \small \textbf{WizardMath} & 13B\\\hline
           \small \textbf{MAmmoTH} & 7B\\\hline
           \small \textbf{MAmmoTH} & 13B \\\hline\hline
        \end{tabular} & 
        \begin{tabular}{r|r|r|r|r|r}
           \multicolumn{6}{c}{\small \textbf{Accuracy }}\\\hline\hline
           \small \textbf{GSM-8K} & \small \textbf{DeepMind} & \small \textbf{NumGLUE} & \small \textbf{SimulEq} & \small \textbf{Math-401*} & \small \textbf{MathQuest} \\\hline\hline
           30.0 & 46.0 & 45.0 & 15.0 & 17.0 & 10.6 \\\hline
           42.0 & 51.0 & 54.0 & 16.0 & 24.0 & 20.3 \\\hline
           64.0 & 55.0 & 52.0 & 29.0 & 15.0 & 16.01 \\\hline
           68.0 & 56.0 & 70.0 & 38.0 & 10.0 & 20.1 \\\hline
           56.0 & 50.0 & 62.0 & 24.0 & 16.0 & 18.5 \\\hline
           67.0 & 51.0 & 64.0 & 34.0 & 18.0 & \textbf{24.0} \\\hline\hline
        \end{tabular}
    \end{tabular}
    \vspace{0.3cm}
    \caption{Exact Match Accuracy Results on the set of 100 samples of 5 datasets and our dataset MathQuest \textbf{After} fine-tuning on Math-401 dataset. (*) refers to the set of Math-401 we augmented for fine-tuning.}
\label{accuracy2}
\end{table}

The dataset was split into training (241,600 samples), validation (30,200 samples), and test (30,200 samples) subsets. We used the AdamW optimizer, a well-recognized technique, to enhance model performance. This optimization step was crucial for achieving the results in our study.

For fine-tuning, we employed QLora~\cite{dettmers2023qlora}, an efficient approach that maximizes memory efficiency and minimize computation cost using 4-bit quantization in a pretrained language model, resulting in Low Rank Adapters (LoRA). Each model underwent 10 epochs of fine-tuning with a learning rate of $3 \times 10^{-4}$. Post fine-tuning, we assessed the models using the same test set employed for pre-fine-tuning inference. The results, summarized in Table.~\ref{accuracy2}, serve to highlight the enhancements achieved in mathematical problem-solving capabilities before and after fine-tuning. 


\subsection{Evaluation Metric}
We compared all model variants to evaluate the quality of the generated solutions. To measure performance, we assessed the accuracy in matching the generated answers to the actual solutions for five open-source datasets: GSM-8K, DeepMind, SimulEq, NumGLUE, and Math-401. These datasets provide ground truth answers for exact match accuracy calculation.

\section{Results \& Discussion\label{results}}
In this section, we present the outcomes of our experiments in the domain of mathematical problem-solving. Our study encompasses evaluations conducted on our proprietary dataset, MathQuest, as well as five other publicly available datasets. This paper establishes baseline performance metrics for the task using our MathQuest dataset. To gauge the effectiveness of Large Language Models (LLMs) across diverse datasets, we utilize exact match accuracy as a benchmark metric. 

We organize our results into two distinct setups: \textbf{before fine-tuning} and \textbf{after fine-tuning} the models, with the primary aim of evaluating the model's learning capabilities. Table.~\ref{accuracy1} presents the exact match accuracy of three models across two variants, 7B and 13B, before fine-tuning, on five datasets and our dataset MathQuest. To summarize these findings, referring to Table.~\ref{accuracy1}, the performance of all the models is notably lower on the SimulEq dataset, as well as on our augmented dataset, Math-401. This discrepancy can be attributed to the presence of intricate problems within these datasets, which often require additional knowledge, such as questions like "Number of red color cards in a deck of 52 cards." Consequently, Table.\ref{accuracy2} provides a detailed overview of the accuracy results following the fine-tuning process. In summary, the accuracy of all models showed significant improvement after undergoing fine-tuning on our diverse and complex question-answer dataset. Notably, models with 13B parameters exhibited higher accuracy compared to those with 7B parameters.

The key takeaways from Table.~\ref{accuracy1}, and Table.~\ref{accuracy2} reveal that the best-performing model is \textbf{MAmmoTH-13B} for our dataset MathQuest, exhibiting the highest accuracy among all models after fine-tuning, at 24.0\%. Additionally, it's noteworthy that both MAmmoTH 7B and 13B generated outputs with precision up to two decimal places, indicating their accuracy. From Table~\ref{accuracy2}, It is evident that our dataset, MathQuest, poses a greater challenge due to its complexity and diversity, resulting in lower accuracy compared to other datasets.

\section{Conclusion\label{conclusion}}
In summary, our approach enhances Large Language Models (LLMs) in acquiring vital reasoning skills for precise mathematical problem-solving. We introduce tailored question-answer pairs in our \textbf{MathQuest} dataset, encompassing single or multiple mathematical operators and expressions. These supportive simple and complex problems guide the model toward incremental problem-solving. Our primary aim is to provide illustrative examples that improve solution accuracy and clarity. Our results demonstrate significant enhancements in both solution precision and comprehensibility, promising valuable support for educators and students seeking effective mathematical problem-solving capabilities.

While our research establishes a robust foundation for advancing mathematical problem-solving with Generative LLMs, further refinements and optimizations are essential to extend its applicability across a broader range of scenarios. Ultimately, our work contributes to advancing conceptual understanding and numerical problem-solving in high school-level mathematical question-answering, offering valuable assistance to students and professionals grappling with complex questions through LLMs.

\section{Limitations}
While our proposed solution can successfully solve basic mathematical problems, it occasionally encounters challenges when dealing with complex mathematical problems that involve retaining variable values for use in subsequent equations.

Another limitation of our proposed work is the partial enhancement of reasoning abilities in LLMs for solving mathematical problems. However, it still falls short in dealing with complex expressions that include nested brackets within equations. The reason could be limited training dataset size, we will try to increase our training data in future research. We intend to address this limitation in our future work, wherein we plan to incorporate recent prompting techniques and further enhance LLMs reasoning abilities for these types of problems.

\section{Acknowledgement}
Dr. Rajiv Ratn Shah is partly supported by the Infosys Center for AI, the Center of Design and New Media, and the Center of Excellence in Healthcare at Indraprastha Institute of Information Technology, Delhi.
We gratefully thank Dr. Astha Verma and Mr. Naman Lal for their guidance and continuous support during our research. Their knowledge and insightful feedback significantly influenced the direction and quality of our research. We appreciate their time, devotion, and willingness to share information, which all contributed considerably to the accomplishment of this job. Their encouragement and constructive talks were a continual source of motivation for us, and we consider ourselves fortunate to have benefited from their wisdom and leadership. 

\bibliographystyle{plain}
\bibliography{neurips_2023}
\end{document}